\documentclass{article}

\usepackage[main, final]{neurips_2025}

\usepackage[utf8]{inputenc} 
\usepackage[T1]{fontenc}    
\usepackage{hyperref}       
\usepackage{url}            
\usepackage{booktabs}       
\usepackage{amsfonts}       
\usepackage{nicefrac}       
\usepackage{microtype}      
\usepackage{xcolor}         

\usepackage{graphicx}
\usepackage{booktabs}
\usepackage{multirow}
\usepackage{adjustbox}
\usepackage{placeins} 
\usepackage{amsmath}

\title{Beyond Single-Sample: Reliable Multi-Sample Distillation for Video Understanding}

\author{%
  Songlin Li \\
  University of Electronic Science and Technology of China \\
  \& Robotics Center, XPeng Motors \\
  \texttt{bigman2243197885@gmail.com} \\
  \And
  Xin Zhu\thanks{Corresponding Author} \\
  Robotics Center, XPeng Motors \\
  \texttt{zhux9@xiaopeng.com} \\
  \AND
  Zechao Guan \\
  Robotics Center, XPeng Motors \\
  \texttt{guanzc2@xiaopeng.com} \\
  \And
  Peipeng Chen \\
  Robotics Center, XPeng Motors \\
  \texttt{chenpp2@xiaopeng.com} \\
  \And
  Jian Yao \\
  Robotics Center, XPeng Motors \\
  \texttt{jian.yao@xiaopeng.com} \\
}

\begin{document}

\maketitle

\begin{abstract}
    Traditional black-box distillation for Large Vision-Language Models (LVLMs) typically 
    relies on a single teacher response per input, which often yields high-variance responses 
    and format inconsistencies in multimodal or temporal scenarios. 
    To mitigate this unreliable supervision, we propose 
    $R\text{-}MSD$ (Reliable Multi-Sample Distillation), a framework that explicitly 
    models teacher sampling variance to enhance distillation stability. Rather than 
    relying on a single teacher response, our approach leverages a task-adaptive 
    teacher pool to provide robust supervision tailored to both closed-ended and 
    open-ended reasoning. By integrating quality-aware signal matching with an
    adversarial distillation objective, our approach effectively filters teacher 
    noise while maximizing knowledge transfer. Extensive evaluations across comprehensive 
    video understanding benchmarks demonstrate that $R\text{-}MSD$ consistently outperforms 
    single sample distillation methods. We additionally include an original SFT+RL
    4B baseline under the same training budget, which shows only marginal gains,
    while our method achieves significant improvements.
    With a 4B student model, our approach delivers gains
    on VideoMME ($+1.5\%$), Video-MMMU ($+3.2\%$), and MathVerse ($+3.6\%$).
    
\end{abstract}

\section{Introduction}
\label{sec:introduction}

Large Vision-Language Models (LVLMs)~\citep{longvila2025,unitime2025,vtgllm2025}
have advanced video understanding, but deployment is hindered by computational costs.
To address this challenge, knowledge distillation~\citep{hinton2015distilling} transfers knowledge from a strong
teacher to a smaller student.
Recent analysis~\citep{yue2025does} shows that distillation can expand a model's reasoning capabilities beyond what reinforcement learning achieves, as RL methods are bounded by the base model's distribution.
While prior work explores multi-teacher aggregation~\citep{you2017learning},
cross-modal transfer~\citep{richkd2025}, and adversarial objectives~\citep{ye2025gad},
these overlook a fundamental question: \emph{is a single teacher sample reliable?}

\begin{figure}[tbp]
    \centering
    \includegraphics[width=1\linewidth]{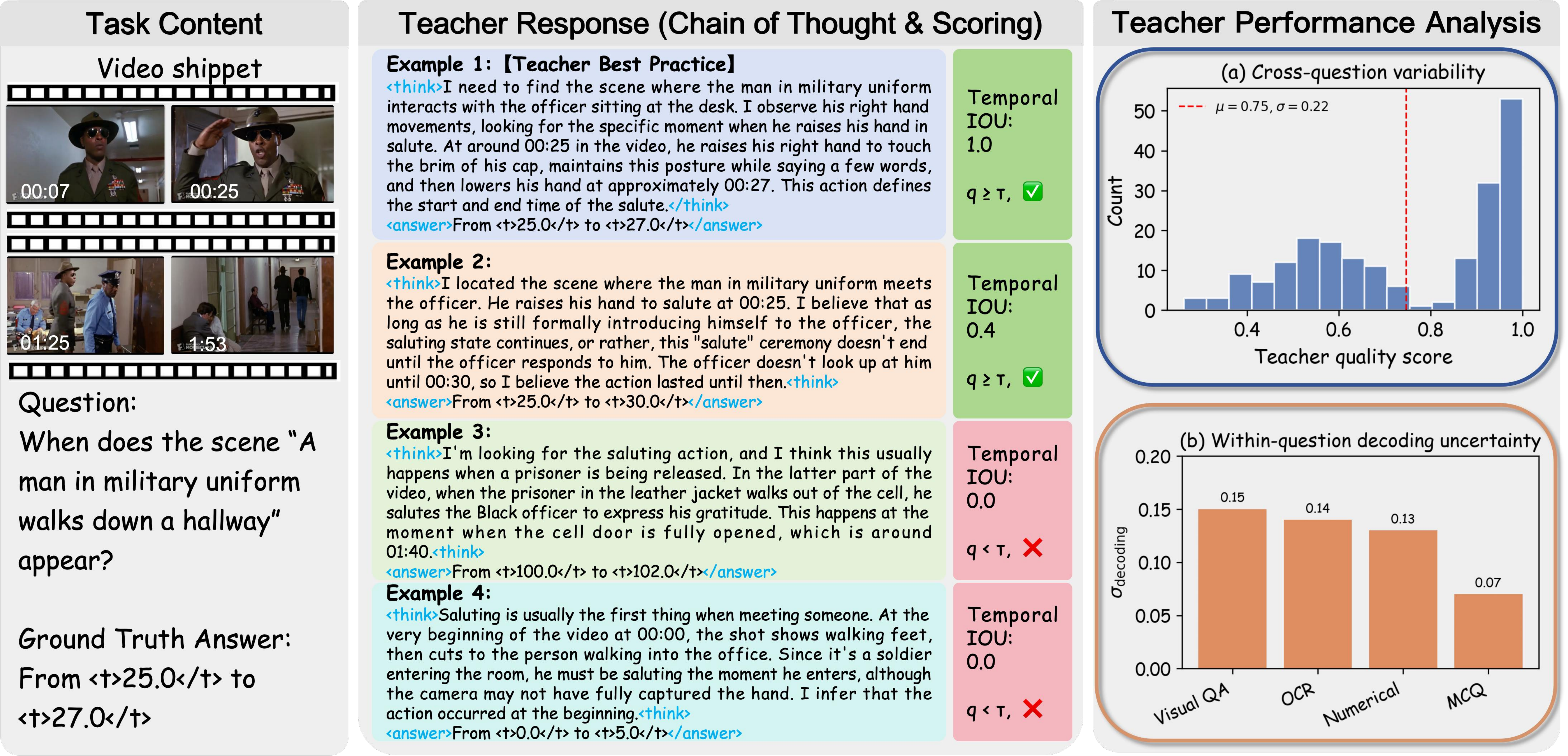}
    \caption{Teacher sampling variance in video understanding.
    \textbf{Left:} Temporal QA example: four teacher responses with their temporal IoU
    to ground truth, showing substantial quality variation across samples of the
    same question.
    \textbf{Right:} Global statistics over 200 teacher responses from eight task types:
    \textbf{(a)} cross-question variance and \textbf{(b)} within-question sampling variance.
    Both dimensions undermine single-sample distillation and motivate our multi-sample approach.}
    \label{fig:two_variance}
\end{figure}

Existing distillation methods---from sequence-level imitation~\citep{kim2016sequence}
to adversarial training~\citep{ye2025gad}---typically assume that a single sampled
teacher response per training input provides reliable supervision.
In text-only settings, this assumption can be acceptable when stochastic decoding
is relatively stable.
In video understanding, we observe a different regime with \emph{two dimensions of variance},
as illustrated in Figure~\ref{fig:two_variance}.
First, \textbf{cross-question variance} is substantial: quality spans $[0.10, 1.0]$
($\mu{=}0.75$, $\sigma{=}0.22$) across 200 samples.
MCQ shows high consistency ($\mu{=}0.96$, $\sigma{=}0.10$), while visual QA is
variable ($\mu{=}0.64$, $\sigma{=}0.24$).
Second, \textbf{within-question uncertainty} is also substantial: repeated sampling
yields $\sigma_{\text{sampling}}$ ranging from $0.07$ (MCQ) to $0.15$ (visual QA),
with quality ranges of $[0.50, 0.85]$ for OCR and $[0.65, 1.00]$ for numerical tasks.
Format violations occur in 1\% of samples overall (10\% for temporal QA).

Teacher-sampling variance interacts with task heterogeneity.
Video benchmarks mix \emph{closed-ended tasks} (verifiable outputs: temporal segments, bounding boxes, MCQ) and \emph{open-ended tasks} (natural language).
Closed-ended tasks allow quality-based filtering via ground-truth; open-ended tasks lack reliable metrics due to lexical-overlap penalties for semantically equivalent responses.
Most distillation pipelines~\citep{dou2025polar} apply uniform supervision and do not model this asymmetry.

To tackle the challenge of unreliable teacher sampling, we propose \textbf{R-MSD}
(\textbf{R}eliable \textbf{M}ulti-\textbf{S}ample \textbf{D}istillation).
The core idea is to sample a \emph{teacher pool} of $K$ responses per input
and use task-adaptive quality signals to \emph{bias which teacher responses are
used for optimization}.
Our training pipeline has two stages: Stage~1 performs SFT warmup on one selected
teacher response per input; Stage~2 performs RL-based adversarial distillation,
where the student generates multiple responses and each rollout is paired with one
teacher response from the pool.
For closed-ended tasks, teacher responses are preferentially paired based on
ground-truth-based quality signals (e.g., temporal/spatial IoU or exact match, together
with format validity checks), so higher-quality teachers are sampled more often.
For open-ended tasks, we use uniform pairing to avoid brittle lexical ranking.
We jointly train a discriminator (critic) on these paired student--teacher
responses to provide distribution-level supervision alongside ground-truth-based rule
rewards.

Our main contributions are:
\begin{itemize}
    \item \textbf{Quantitative analysis of teacher decoding uncertainty.}
    We identify \emph{teacher decoding uncertainty} as a central source of supervision
    noise in video LVLM distillation. Across 200 samples, quality varies substantially
    ($\sigma{=}0.22$ globally, up to $\sigma{=}0.29$ by task), and format violations are non-negligible
    (1.0\% overall, 3.5\% task-specific), showing that single-sample supervision is unreliable.
    
    \item \textbf{Method: R-MSD (Reliable Multi-Sample Distillation).}
    We propose a two-stage framework with a teacher pool per input and
    task-adaptive matching: quality-biased pairing for closed-ended tasks and
    uniform pairing for open-ended tasks. Combined with an online
    critic-as-discriminator, this improves supervision \emph{quality}, not just
    supervision \emph{quantity}.

    \item \textbf{Comprehensive validation.}
    We evaluate R-MSD on six video and two image QA benchmarks, demonstrating
    strong cross-domain transfer. Under the same test-time frame count, our
    distilled 4B student surpasses Qwen3-VL-4B on VideoMME (65.3 vs.\ 63.8),
    Video-MMMU (58.6 vs.\ 55.4), and WorldSense (49.2 vs.\ 46.7).
    An SFT+RL 4B baseline trained under a matched protocol yields only marginal
    gains compared with our task-adaptive multi-sample distillation.
\end{itemize}

\section{Related Work}
\label{sec:related}

\subsection{Video Understanding with Vision-Language Models}
In the video domain, image LVLMs are extended with video-aware encoders,
temporal tokenization, and video-instruction data.
Representative systems include VideoChat/VideoChatGPT for conversational
reasoning~\citep{li2023videochat,maaz2023videochatgpt}, Video-LLaVA for unified
image--video representations~\citep{lin2023videollava}, and efficiency-oriented
adaptations such as LLaMA-VID and PLLaVA~\citep{li2023llamavid,xu2024pllava}.
Recent pipelines also improve data scale/quality via synthetic instruction and
captioning refinement~\citep{zhang2025llavavideo,chen2024sharegpt4video}.

The central open issue is no longer whether LVLMs can follow video instructions,
but how to optimize generation quality when outputs are open-ended and
evaluation signals are noisy.
Although supervised fine-tuning (SFT) is a strong baseline, open-ended video
understanding often requires optimization beyond maximum-likelihood training.
This motivates preference learning and reinforcement learning (RL) for direct
quality optimization.
In language modeling, RLHF~\citep{ouyang2022training} and direct alignment methods such as DPO~\citep{rafailov2023dpo} have become standard tools for quality optimization.
Analogously, multimodal alignment has explored using large (multi)modal models as judges or feedback providers~\citep{wang2024rlvlmf,zhu2024nsft}.
However, reward design for open-ended answers remains a central bottleneck: scalar rewards can be noisy, and learned reward models may be exploited by the policy during optimization (reward hacking)~\citep{gao2022scaling}.
Recent analysis~\citep{yue2025does} shows that reinforcement learning with verifiable rewards (RLVR) improves sampling efficiency but does not expand reasoning capabilities beyond the base model's distribution---whereas distillation can transfer genuinely new reasoning patterns from a stronger teacher.

Our work addresses a complementary reliability issue that appears \emph{before}
reward modeling: with few sampled teacher responses, the supervision signal
itself can be unstable.
Instead of relying on a fixed, offline-trained multimodal reward model as a
judge (which can be vulnerable to reward hacking as the policy improves), we
adopt an \emph{adversarial distillation} perspective where the reward/critic is
trained \emph{online} alongside the student.
This co-evolution provides adaptive supervision for open-ended video responses
and reduces reliance on a static reward proxy that may become brittle as the
policy improves.

\subsection{Knowledge Distillation for Sequence Generation}
Knowledge distillation (KD) transfers knowledge from a large teacher to a
smaller student.
Classic KD matches softened logits~\citep{hinton2015distilling},
while modern compression techniques have been widely applied to Transformer architectures.
For sequence generation, sequence-level distillation trains a student on
teacher-generated outputs to reduce exposure bias~\citep{kim2016sequence}.
With the rise of instruction-tuned LLMs, distillation has become a standard approach for creating smaller, capable assistants.

A key practical challenge is that many high-performing teachers are
\emph{black boxes} (accessible only via an API), making logit-based KD infeasible.
This motivates response-based distillation methods that learn only from sampled teacher responses.
Recently, Generative Adversarial Distillation (GAD) formulates black-box,
on-policy distillation as a minimax game where a discriminator distinguishes
student outputs from teacher outputs and serves as an online reward
signal~\citep{ye2025gad}.
Concurrently, reward modeling has been re-interpreted as training policy discriminators, e.g., POLAR, which learns a general reward model by discriminating policies rather than predicting absolute preference scores~\citep{dou2025polar}.
Recent OPD analysis further connects on-policy distillation to KL-constrained RL and studies reward-scaling and reference-model choices in a generalized formulation~\citep{yang2026learningteachergeneralizedonpolicy}.
In this paper, we use \emph{on-policy distillation} to denote this distillation paradigm in prior work (online sample generation with discriminator/reward co-training), which is distinct from the RL taxonomy term \emph{on-policy RL}.

\paragraph{Two complementary trends in response-based distillation.}
Existing methods can be organized into two trends.
The first is \emph{supervised-style distillation}: sequence-level imitation,
multi-sample filtering, and quality-aware selection
~\citep{kim2016sequence,wang2022eekd,zhou2023lima,dong2023raft}.
The second is \emph{RL/adversarial-style distillation}: on-policy
discriminator/reward training for distribution-level alignment
~\citep{ye2025gad,dou2025polar,yang2026learningteachergeneralizedonpolicy}.
R-MSD unifies these trends: it extends supervised-style distillation with task-adaptive quality weighting, while preserving online distribution-level alignment via adversarial objectives.

\paragraph{Multi-sample and quality-aware distillation.}
A natural extension of single-sample distillation is to use multiple teacher
outputs per input.
Multi-teacher distillation aggregates knowledge from diverse teachers~
\citep{you2017learning,fukuda2017efficient}, while self-ensemble methods generate multiple outputs from a single teacher to improve robustness~
\citep{wang2022eekd}.
Recent work has also explored quality-aware selection strategies, such as filtering low-quality samples based on confidence scores~
\citep{zhou2023lima} or using reward models to rank teacher outputs~
\citep{dong2023raft}.
However, these approaches are primarily developed for text-only tasks and rely
on logit-based signals or text-specific quality metrics.
In the video domain, quality assessment must account for closed-ended outputs
(e.g., temporal segments) and the unreliability of lexical metrics for open-ended
responses---a distinction largely overlooked by prior work.
While these advances are promising, existing black-box adversarial distillation
methods are primarily developed and evaluated in \emph{pure-text} settings.
Extending them to video LVLMs is challenging due to (i) a larger and noisier
observation space, (ii) sensitivity to temporal sampling and grounding, and
(iii) higher output-format heterogeneity across benchmarks.
Multi-sample and multi-teacher distillation has been explored to use diverse
teachers or generations, but many approaches emphasize teacher diversity rather
than the stability of sampled supervision from a single strong teacher.

Our method builds on GAD-style online adversarial distillation and extends it
to video-specific supervision noise.
The key difference is explicit modeling of teacher decoding uncertainty with
task-adaptive matching: ground-truth-informed quality estimation for closed-ended tasks and
uniform treatment for open-ended supervision.
This design reduces reliance on brittle lexical proxies while preserving online
distribution-level alignment through an adaptive discriminator.

Overall, prior work leaves two gaps most relevant to our setting:
(i) robust quality handling for open-ended video responses, and
(ii) explicit modeling of supervision noise induced by teacher sampling.
Our framework targets both by coupling R-MSD-style task-adaptive multi-sample supervision
with online adversarial alignment.

\section{Method}
\label{sec:method}

\subsection{Problem Formulation}
Given a video-question-answer tuple $(V, Q, y^*)$, our goal is to train a student
model $\pi_S$ from a stronger teacher $\pi_T$ under black-box access.
Unlike conventional sequence-level distillation that uses a single teacher sample,
we model teacher sampling variance by drawing $K$ teacher responses
for the same input:
\begin{equation}
\{T_1, \ldots, T_K\} \sim \pi_T(\cdot \mid V, Q).
\label{eq:teacher_pool}
\end{equation}

Our objective is to convert this multi-sample set into a stable supervision
signal that emphasizes reliable responses when correctness is verifiable and
avoids lexical bias when correctness is not reliably measurable.

\subsection{Overall Framework}
\label{subsec:framework}
\begin{figure}[tbp]
    \centering
    \includegraphics[width=0.95\linewidth]{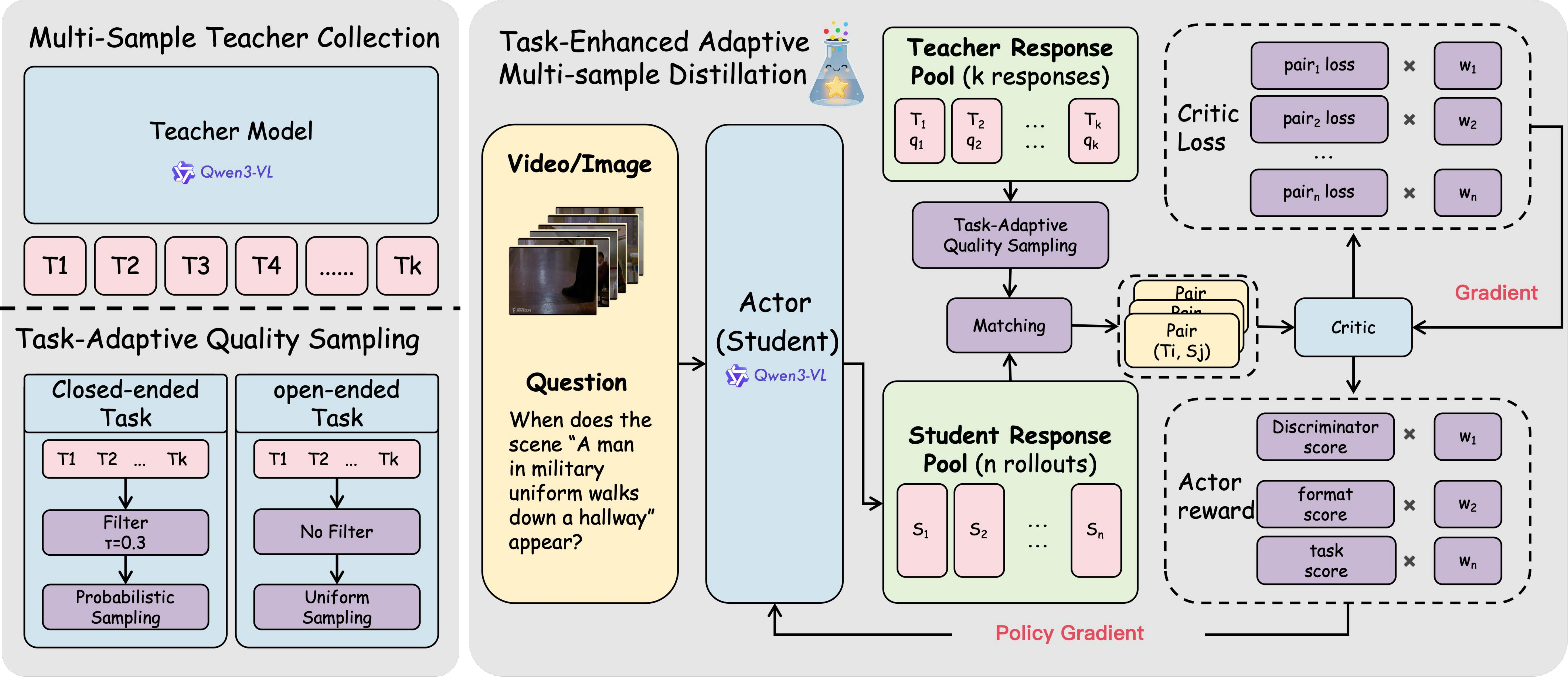}
    \caption{Overview of the R-MSD framework (Stage 2: RL-based 
    Adversarial Distillation). The pipeline utilizes a multi-sample teacher 
    collection to provide diverse references. A critic-as-discri\-minator 
    mechanism is employed to perform task-adaptive sampling, matching 
    student rollouts with teacher responses. The student policy is 
    then iteratively refined through adversarial distillation, 
    using a weighted combination of discriminator, format, and 
    task-specific scores as the reward signal.}
    \label{fig:framework}
\end{figure}

Figure~\ref{fig:framework} illustrates \textbf{R-MSD} (Reliable Multi-Sample Distillation), which has three components: (1) \emph{multi-sample teacher collection} (sample $K$ teacher outputs per input), (2) \emph{task-adaptive quality assessment} (GT-based scoring for closed-ended tasks, uniform weighting for open-ended tasks), and (3) \emph{task-adaptive matching with online discrimination} (quality-aware pairing with a critic-as-discri\-minator).
Training proceeds in two stages: \textbf{Stage 1} (SFT warmup) and \textbf{Stage 2} (RL-based adversarial distillation).

\subsection{Task-Adaptive Quality Assessment and Matching}
\label{subsec:quality_assessment}

\paragraph{Task taxonomy.}
We classify tasks based on answer verifiability.
\emph{Closed-ended tasks} have answers that can be objectively verified against ground-truth annotations (e.g., temporal/spatial grounding, multiple-choice, binary QA, numerical regression, OCR).
\emph{Open-ended tasks} require open-ended natural language descriptions (e.g., detailed visual event description) where no generally reliable correctness metric exists.

\subsubsection{Quality scoring for closed-ended tasks}
For closed-ended tasks, we assign each teacher response $T_k$ a ground-truth-based quality
score $q_k \in [0,1]$:
\begin{equation}
q_k = \mathbb{I}(\text{valid}(T_k)) \cdot \text{Metric}(T_k, y^*),
\label{eq:quality_score}
\end{equation}
where $\mathbb{I}(\text{valid}(T_k))$ indicates format validity and $\text{Metric}(\cdot)$ is the task-specific metric.
For MCQ tasks, the metric uses exact match: $q_k = 1$ if the predicted answer option matches the ground-truth option, and $q_k = 0$ otherwise.
For temporal grounding we use temporal IoU; for spatial grounding we use IoU; for numerical tasks we use $\epsilon$-accuracy.
Unparseable or format-invalid responses receive $q_k=0$.
Samples with $q_k=0$ remain in the teacher pool but have $p_k=0$, meaning they are never selected during matching.
For closed-ended tasks, we optionally apply quality filtering by setting $p_k = 0$ for teacher samples with $q_k < \tau$, where $\tau$ is a quality threshold (we use $\tau = 0.3$ in experiments, validated via sensitivity analysis in Table~\ref{tab:sensitivity}).

\subsubsection{Uniform treatment for open-ended tasks}
For open-ended tasks, we avoid lexical-overlap-based ranking by default to prevent false negatives from synonym/paraphrase variations.
We therefore use uniform matching for this task family, setting $p_k = 1/K$ in Eq.~\ref{eq:matching}.

\subsubsection{Quality-weighted matching}
Our implementation realizes ``quality weighting'' in Stage~2 via
\emph{quality-weighted matching} between multiple student rollouts and a teacher
pool.
For each input $(V,Q)$, we maintain a teacher pool $\{T_1,\dots,T_K\}$.
During on-policy optimization, the student samples $N$ responses
$\{S_1,\dots,S_N\}$ for the same prompt.
We then match each $S_i$ to one teacher response $T_{m(i)}$:
\begin{equation}
    m(i) \sim \text{Categorical}(p_1,\dots,p_K),
    \label{eq:matching}
\end{equation}
where the sampling distribution is task-adaptive.
For closed-ended tasks,
\begin{equation}
    p_k = \frac{q_k}{\sum_{j=1}^{K} q_j},
    \label{eq:pk_closed}
\end{equation}
after applying quality filtering ($q_k = 0$ for filtered samples).
Higher-quality teacher samples are more likely to be paired and may be selected multiple times.
For open-ended tasks, we use uniform matching $p_k = 1/K$.
The matching probability $p_k$ determines which teacher response $T_{m(i)}$ supervises each student rollout $S_i$ in Stage~2.
This avoids two failure modes: treating all closed-ended responses equally despite quality gaps, and over-penalizing open-ended responses with unreliable lexical metrics.
Consequently, our method improves the reliability of supervision rather than merely increasing the number of training samples.

\subsection{Two-Stage Training Objective}
\label{subsec:two_stage}

\paragraph{Stage 1: SFT warmup (single-teacher selection).}
We initialize the student $\pi_S$ with supervised fine-tuning on a single
teacher response per input.
When multiple teacher responses are available in the dataset, we select one
response per input: for closed-ended tasks, we select the response with the
highest quality score $q_k$; for open-ended tasks, we randomly select one
response from the pool to avoid introducing bias from arbitrary selection criteria.
We then apply standard autoregressive cross-entropy:
\begin{equation}
\mathcal{L}_{\text{SFT}} = - \log \pi_S(T_{\text{best}} \mid V, Q).
\label{eq:sft_loss}
\end{equation}
This stage provides stable initialization before adversarial training.

\paragraph{Stage 2: RL-based adversarial distillation with composite rewards.}
We optimize the student using policy-gradient RL on freshly sampled student rollouts with a composite reward that combines distribution-level and content-level signals.
For each prompt, we sample $N$ student responses $S_i \sim \pi_S(\cdot\mid V,Q)$, each paired with a teacher response $T_{m(i)}$ via task-adaptive matching.
Eq.~\ref{eq:reward} decomposes the reward into four components:
\begin{equation}
R(S_i) = \alpha\,D_\phi(S_i) + \beta\,R_{\text{outer}}(S_i) + \eta\,R_{\text{task}}(S_i) + \delta\,R_{\text{content}}(S_i),
\label{eq:reward}
\end{equation}
where the weights satisfy $\alpha + \beta + \eta + \delta = 1$ to maintain reward scale consistency, $D_\phi(S_i)$ is the discriminator score (higher values indicate more teacher-like outputs),
$R_{\text{outer}}$ encourages a valid outer response format, $R_{\text{task}}$
checks task-specific answer formatting (e.g., timestamps / options / boxes), and
$R_{\text{content}}$ measures correctness for closed-ended tasks.
\paragraph{Reward specification.}
$R_{\text{outer}} \in \{0, 1\}$ indicates valid outer response format (1 if response contains valid tags including `<answer>'` and thinking tags like ``);
$R_{\text{task}} \in \{0, 1\}$ checks task-specific format compliance (e.g., temporal grounding with `<t>x</t>` to `<t>x</t>`, MCQ option letters);
$R_{\text{content}} \in [0, 1]$ is the GT-based quality score identical to $q_k$ in Eq.~\ref{eq:quality_score} for closed-ended tasks (0 for open-ended).
By separating format validity from content correctness, we align reward shaping with evaluation metrics rather than relying on brittle lexical overlap proxies.

\subsection{Discriminator and Adversarial Training}
\label{subsec:discriminator}

\paragraph{Discriminator architecture.}
We implement the discriminator as a \emph{critic-as-discri\-minator} that assigns a
scalar score to a generated response.
Concretely, we reuse the critic value head to score the last token of a
response, yielding a single scalar per sequence.
Higher scores indicate teacher-like responses.

\paragraph{Discriminator loss.}
Given a matched student--teacher pair $(S_i, T_{m(i)})$, the discriminator is
trained with the quality-weighted GAD pairwise objective:
\begin{equation}
\mathcal{L}_{D} = \mathbb{E}_{i}\Big[q_{m(i)} \cdot -\log \sigma\big(D_\phi(T_{m(i)}) - D_\phi(S_i)\big)\Big],
\label{eq:disc_loss}
\end{equation}
where $\sigma$ is the sigmoid function and $q_{m(i)}$ is the quality score of teacher response $T_{m(i)}$ (Eq.~\ref{eq:quality_score}).
Task-adaptive quality weighting enters through the matching distribution
$m(i)$ (closed-ended tasks preferentially pair higher-quality teachers), rather
than via an explicit per-sample weight inside $\mathcal{L}_{D}$.

\paragraph{Student RL objective.}
The student is trained to maximize the composite reward from Eq.~\ref{eq:reward}.
Using the policy gradient formulation, the complete Stage 2 objective is:
\begin{equation}
\mathcal{L}_{\text{RL}} = -\mathbb{E}_{S \sim \pi_S}\big[R(S)\big] + \gamma\,D_{\text{KL}}(\pi_S \| \pi_{\text{ref}}),
\label{eq:rl_objective}
\end{equation}
where $R(S)$ is the composite reward from Eq.~\ref{eq:reward} and the KL penalty term prevents the policy from deviating too far from a reference policy (initialized from Stage 1).
The adversarial component corresponds to the discriminator term in $R(S)$, which encourages the student to produce outputs that receive high discriminator scores (i.e., teacher-like outputs).

\subsection{Complexity and Practical Considerations}

Relative to single-sample distillation, our method introduces additional teacher
sampling cost proportional to $K$. In exchange, training uses richer supervision
without changing the inference-time architecture. Thus, the additional cost is
paid only during distillation training.
In practice, $K$ controls a direct quality--cost trade-off, which we quantify in
the sensitivity study in Table~\ref{tab:sensitivity}.

\section{Experiments}
\label{sec:experiments}

\begin{table*}[t]
\centering
\caption{Main results on video and image QA benchmarks. Metrics follow official protocols (accuracy for QA and $\Delta_{\text{knowledge}}$ for Video-MMMU). \textbf{Frames} denotes test-time frames. Under the same 4B scale and frame budget, R-MSD outperforms both the base model and an original SFT+RL baseline.}
\label{tab:main_results}
\small
\begin{adjustbox}{width=\textwidth,center}
\begin{tabular}{lcccccccc}
\toprule
\multirow{2}{*}{Model} & \multirow{2}{*}{Frames} & \multicolumn{5}{c}{Video QA} & \multicolumn{2}{c}{Image QA} \\
\cmidrule(lr){3-7} \cmidrule(lr){8-9}
& & VideoMME & Video-MMMU & WorldSense & LongVideoBench & MLVU\_MCQ & MathVista & MathVerse \\
\midrule
\multicolumn{9}{l}{\textit{Proprietary Models}} \\
GPT-4o~\citep{openai2024gpt4o} & - & 71.9 & 61.2 & - & 66.7 & - & 63.8 & 41.2 \\
Gemini-2.5-Pro~\citep{google2025gemini} & - & 84.3 & 83.6 & - & 77.6 & - & 82.7 & - \\
Seed1.5-VL~\citep{seed2025seed15} & - & 77.9 & 81.4 & - & 74.0 & 82.1 & 85.6 & - \\
\midrule
\multicolumn{9}{l}{\textit{2B-Scale Open Models}} \\
InternVL2.5-2B~\citep{internvl2024} & - & 51.9 & - & - & 52.0 & 61.4 & 51.3 & 30.6 \\
InternVL3-2B~\citep{internvl2025} & - & 58.9 & - & - & 55.4 & 64.2 & 57.0 & 25.3 \\
InternVL3.5-2B~\citep{internvl2025a} & - & 58.4  & - & - & 57.4 & 64.4 & 71.8 & 53.4 \\
VideoLLaMA3-2B~\citep{videollama2025} & - & 59.6 & - & - & 57.1 & 65.4 & 59.2 & - \\
Qwen3-VL-2B-Instruct~\citep{qwen2025qwen3vl} & 64 & 60.8 & 39.5 & 42.6 & 53.6 & 66.8 & 58.9 & 51.1 \\
\midrule
\multicolumn{9}{l}{\textit{4B-Scale Open Models}} \\
InternVL2.5-4B~\citep{internvl2024} & - & 62.3 & - & - & 55.2 & 68.3 & 60.5 & 37.1 \\
InternVL3.5-4B~\citep{internvl2025a} & - & 65.4  & - & - & 60.8 & 70.4 & 77.1 & 61.7 \\
Qwen3-VL-4B-Instruct~\citep{qwen2025qwen3vl} & 64 & 63.8 & 55.4 & 46.7 & 59.3 & 72.4 & 69.5 & 45.7 \\
Original SFT+RL (4B) & 64 & 64.0 & 55.9 & 46.3 & 57.2 & 73.1 & 71.2 & 46.8 \\
\midrule
\multicolumn{9}{l}{\textit{7B-8B-Scale Open Models}} \\
InternVL2.5-8B~\citep{internvl2024} & - & 64.2 & - & - & 60.0 & 68.9 & 64.4 & 39.5 \\
InternVL3-8B~\citep{internvl2025} & - & 66.3 & - & - & 58.8 & 71.4 & 71.6 & 39.8 \\
InternVL3.5-8B~\citep{internvl2025a} & - & 66.0 & - & - & 62.5 & 70.2 & 78.4 & 61.5 \\
VideoLLaMA3-7B~\citep{videollama2025} & - & 66.2 & - & - & 59.8 & 44.1 & 67.1 & - \\
Qwen3-VL-8B-Instruct~\citep{qwen2025qwen3vl} & 64 & 64.0 & 63.3 & - & 61.5 & 75.1 & 74.2 & 58.1 \\
\midrule
\multicolumn{9}{l}{\textit{Ours}} \\
\textbf{Ours (2B)} & 64 & \textbf{61.9} & \textbf{40.8} & \textbf{45.7} & \textbf{53.7} & \textbf{68.2} & \textbf{61.4} & \textbf{55.3} \\
\textbf{Ours (4B)} & 64 & \textbf{65.3} & \textbf{58.6} & \textbf{49.2} & \textbf{58.8} & \textbf{73.2} & \textbf{72.1} & \textbf{49.3} \\
\bottomrule
\end{tabular}
\end{adjustbox}
\end{table*}

\begin{table*}[t]
\centering
\caption{Results on V-STaR~\citep{cheng2025vstar}. We report When (temporal IoU) and Where (visual IoU) under the official evaluation. \textbf{Frames} denotes test-time frames. Improvements indicate better spatio-temporal grounding under teacher sampling variance.}
\label{tab:vstar_results}
\small
\renewcommand{\arraystretch}{0.95}
\begin{tabular}{lccccc}
\toprule
Model & Frames & \multicolumn{2}{c}{When (Temporal IoU)} & \multicolumn{2}{c}{Where (Visual IoU)} \\
\cmidrule(lr){3-4} \cmidrule(lr){5-6}
 &  & Chain1 & Chain2 & Chain1 & Chain2 \\
\midrule
GPT-4o~\citep{openai2024gpt4o} & - & 16.7 & 12.8 & 6.5 & 3.0 \\
Gemini-2-Flash~\citep{google2025gemini} & - & 24.5 & 23.8 & 4.6 & 2.2 \\
Video-LLaMA3~\citep{videollama2025} & - & 23.0 & 23.1 & 0.9 & 0.2 \\
LLaVA-Video~\citep{zhang2025llavavideo} & - & 10.5 & 12.2 & 1.9 & 1.3 \\
VideoChat2~\citep{li2024mvbench} & - & 13.7 & 12.5 & 2.5 & 1.0 \\
Oryx-1.5-7B~\citep{liu2024oryx} & - & 13.5 & 14.8 & 10.1 & 3.5 \\
InternVL-2.5-8B~\citep{internvl2024} & - & 8.7 & 7.8 & 0.7 & 0.1 \\
Qwen2.5-VL-7B~\citep{qwen2025vl} & - & 15.4 & 13.8 & 17.0 & 2.5 \\
TRACE~\citep{guo2025trace} & - & 19.1 & 17.1 & 0.0 & 0.0 \\
Sa2VA-8B~\citep{yuan2025sa2va} & - & 0.1 & 0.0 & 32.3 & 37.5 \\
Open-o3 Video~\citep{chen2025openo3} & - & 24.5 & 24.0 & 25.4 & 6.0 \\
Qwen3-VL-4B~\citep{qwen2025qwen3vl} & 64 & 21.3 & 18.5 & 22.3 & 5.0 \\
\textbf{Ours (4B)} & 64 & \textbf{25.2} & \textbf{23.4} & \textbf{24.8} & \textbf{7.0} \\
\bottomrule
\end{tabular}
\end{table*}

We design experiments to evaluate three claims from
section~\ref{sec:introduction}: (1) R-MSD improves over single-sample baselines under matched budgets,
(2) each component contributes to the final gain,
and (3) teacher sampling variance is a practical bottleneck.
The section is organized accordingly:
subsection~\ref{subsec:main_results} focuses on Claim~1,
subsection~\ref{subsec:ablation} examines Claim~2, and
subsection~\ref{subsec:variance_analysis} provides dedicated evidence for Claim~3.

\subsection{Experimental Setup}
\label{subsec:exp_setup}

\subsubsection{Benchmarks}
We evaluate on six public video understanding benchmarks : VideoMME, Video-MMMU, WorldSense, LongVideoBench, MLVU\_MCQ, and VsTAR~\citep{cheng2025vstar},
plus two image QA benchmarks (MathVista, MathVerse) for cross-domain evaluation.
All benchmarks use official evaluation splits and metrics.
VideoMME, WorldSense, LongVideoBench, and MLVU\_MCQ use multiple-choice accuracy; Video-MMMU uses the knowledge-gain metric $\Delta_{\text{knowledge}}$;
VsTAR uses grounding metrics (tIoU/IoU); MathVista and MathVerse use accuracy.
With one exception (VsTAR, evaluated via tIoU/IoU), all benchmarks are treated as closed-ended tasks with exact-match evaluation.

\subsubsection{Training data}
We construct our distillation dataset from VideoR1~\citep{wang2024videor1} (10\% subset) and Open-O3~\citep{chen2025openo3} (full dataset).
The frozen Qwen3-VL-235B teacher processes raw video--question pairs to generate supervision annotations.
Stage~1 SFT warmup uses 50K samples; Stage~2 RL-based adversarial distillation optimization uses 60K samples with multi-sample quality signals.
All training samples are disjoint from evaluation benchmarks.

\subsubsection{Models and training protocol}
We use a frozen Qwen3-VL-235B teacher and Qwen3-VL-4B student under black-box distillation.
We set teacher pool size $K{=}4$, sample $N{=}8$ student rollouts per input during Stage~2, and train for 1 epoch per stage.
Optimization uses AdamW (learning rate $2\times 10^{-6}$ for student, $1\times 10^{-6}$ for discriminator) with batch size 128.
The reward weights are $(\alpha, \beta, \eta, \delta) = (0.4, 0.1, 0.1, 0.4)$.
Students train with 16 frames per video and are evaluated with 64 frames at test time.

\subsection{Main Results}
\label{subsec:main_results}

After introducing the setup, we first evaluate Claim~1 by comparing our distilled 4B student with
existing video LVLMs on the full suite of six video benchmarks, and also
examines cross-domain transfer on two image QA benchmarks.
All evaluations use each benchmark's official split and metric.

\paragraph{Compared models.}
We compare against publicly available video LVLMs at similar scale.
Larger models are included only as context, not direct baselines.
Our 4B student is distilled from a larger frozen teacher using R-MSD
(section~\ref{sec:method}), and prior-work numbers use official checkpoints/scripts when available.
We additionally report an in-house \emph{original SFT+RL} 4B baseline under the same
teacher, student scale, training budget, and frame budget for direct protocol comparison.

\paragraph{Results.}
Table~\ref{tab:main_results} summarizes five video QA and two image QA benchmarks; Table~\ref{tab:vstar_results} reports VsTAR separately.
Under the same 64-frame test budget, our 4B student consistently outperforms Qwen3-VL-4B.
Gains range from +0.8 (MLVU) to +3.6 (MathVerse), with VideoMME (+1.5), Video-MMMU (+3.2), WorldSense (+2.5), and MathVista (+2.6); LongVideoBench shows no significant change.
On VsTAR (Table~\ref{tab:vstar_results}), our 4B student also improves on all four chains, with Chain2 temporal IoU (+4.9) and Chain2 visual IoU (+2.0).

\paragraph{Interpretation.}
Consistent improvements confirm R-MSD enhances supervision reliability; non-uniform gains align with teacher sampling variance across tasks.

\subsection{Ablation and Analysis}
\label{subsec:ablation}

This subsection evaluates Claim~2 by dissecting which components of R-MSD are
responsible for the gains beyond multi-sample training and how sensitive the
method is to the number and quality of teacher samples.

\subsubsection{Core component ablation}
We compare four settings to isolate gains beyond naive multi-sample training: A ($K{=}1$), B ($K{=}4$), C (+filtering, $\tau{=}0.3$), and D (+weighting).  
Except A, all settings use $K{=}4$ with the same teacher, student, and training schedule.  
\textbf{Filtering} removes low-quality teacher samples ($q_k < \tau$) for closed-ended tasks.  
\textbf{Weighting} trains the discriminator with quality weights, giving more influence to higher-quality teacher responses.  
Table~\ref{tab:core_ablation} reports results on VideoMME and Video-MMMU.

\begin{table}[t]
\centering
\small
\caption{Core component ablation on VideoMME and Video-MMMU (accuracy, \%). $K$: teacher sample count. ``Filter'' applies threshold $\tau$ to remove low-quality teacher samples ($q_k < \tau$). ``Weight'' enables quality-weighted discriminator updates by $q_k$. Setting A reports Stage~1 single-sample SFT; vanilla Qwen3-VL-4B is in Table~\ref{tab:main_results}.}
\label{tab:core_ablation}
\begingroup
\renewcommand{\arraystretch}{0.95}
\begin{tabular}{lccccc}
\toprule
Setting & $K$ & Filter & Weight & VideoMME & Video-MMMU \\
\midrule
A (single-sample) & 1 & No & No & 63.8 & 54.4 \\
B (multi-sample) & 4 & No & No & 64.5 & 55.9 \\
C (+filtering) & 4 & Yes & No & 65.0 & 57.2 \\
D (ours) & 4 & Yes & Yes & 65.3 & 58.6 \\
\bottomrule
\end{tabular}
\endgroup
\end{table}

\begin{table}[t]
\centering
\small
\caption{Sensitivity analyses for teacher sample count $K$ (left; VideoMME/Video-MMMU accuracy, \%) and quality threshold $\tau$ (right; valid-sample ratio and accuracy) under the full method. We select $\tau=0.3$ to balance low-quality removal (72\% retention) and teacher-pool diversity.}
\label{tab:sensitivity}
\begingroup
\renewcommand{\arraystretch}{0.95}
\begin{tabular}{ccc}
\toprule
$K$ & VideoMME & Video-MMMU \\
\midrule
2 & 64.8 & 57.1 \\
4 & 65.3 & 58.6 \\
8 & 65.5 & 58.9 \\
\bottomrule
\end{tabular}
\hspace{1.8em}
\begin{tabular}{cccc}
\toprule
$\tau$ & Valid-sample ratio & VideoMME & Video-MMMU \\
\midrule
0.0 & 100\% & 64.5 & 55.9 \\
0.2 & 87\% & 65.0 & 58.1 \\
0.3 & 72\% & 65.3 & 58.6 \\
0.5 & 45\% & 64.8 & 57.2 \\
\bottomrule
\end{tabular}
\endgroup
\end{table}

\subsubsection{Sensitivity to sample count $K$}
Table~\ref{tab:sensitivity} (left) shows that increasing $K$ from $1$ to $4$ improves both benchmarks, while $K{=}8$ yields only marginal extra gain.

\subsubsection{Sensitivity to quality threshold $\tau$}
Table~\ref{tab:sensitivity} (right) shows that moderate filtering ($\tau{=}0.2$ or $0.3$) improves both VideoMME and Video-MMMU, whereas aggressive filtering ($\tau{=}0.5$) removes too many samples and slightly hurts performance.
This supports discarding clearly low-quality teacher responses while preserving teacher-pool diversity.

\subsubsection{Pass@k on Video-MMMU}
Following recent work on evaluating reasoning capability boundaries~\citep{yue2025does}, we use Pass@$k$ to measure reliability under repeated decoding: for each Video-MMMU query, we sample $k$ responses with $T{=}1.0$ and top-$p{=}0.9$, counting a success if any response matches the ground-truth option.
Unlike pass@1 which reflects average-case behavior, pass@$k$ at large values reveals a model's problem-solving potential~\citep{yue2025does}.

Figure~\ref{fig:variance_and_passk} (right) shows the Pass@$k$ curve from $k{=}1$ to $k{=}128$.
Our method achieves +3.2\% higher Pass@1 than Qwen3-VL-4B while converging to a similar upper bound, indicating R-MSD primarily improves the probability of correct answers with fewer inference calls.

\subsubsection{Task-adaptive strategy validation}
Table~\ref{tab:task_adaptive} compares GT-based scoring versus uniform weighting across closed-ended and open-ended task subsets.

\begin{table}[t]
\centering
\caption{Task-adaptive strategy validation: average accuracy (\%) over closed-ended and open-ended task subsets across VideoMME and Video-MMMU under GT-based scoring vs.\ uniform weighting.}
\label{tab:task_adaptive}
\small
\begingroup
\renewcommand{\arraystretch}{0.95}
\begin{tabular}{lcc}
\toprule
Task family & GT-based scoring & Uniform weighting \\
\midrule
Closed-ended tasks & 57.8 & 56.2 \\
Open-ended tasks & 58.4 & 59.1 \\
\bottomrule
\end{tabular}
\endgroup
\end{table}
For closed-ended tasks, GT-based scoring outperforms uniform weighting, while for open-ended tasks the trend reverses.
This supports our task-adaptive design: lexical metrics for open-ended responses can hurt performance.

\paragraph{Reproducibility.}
We follow official benchmark evaluation scripts; when prior work lacks frame counts or evaluation details, we retain published numbers as context without direct fairness claims beyond matched settings.

\subsection{Teacher Sampling Variance Analysis}
\label{subsec:variance_analysis}

The main results in subsection~\ref{subsec:main_results} show that R-MSD achieves consistent
improvements across benchmarks, but with non-uniform gain patterns (e.g., +3.2 on
Video-MMMU vs.\ +0.8 on MLVU\_MCQ). To understand why improvements vary by task, we
analyze teacher sampling variance and examine how it relates to the performance
gains we observe.

\begin{figure}[htbp]
    \centering
    \begin{minipage}[t]{0.48\linewidth}
        \centering
        \includegraphics[width=\linewidth]{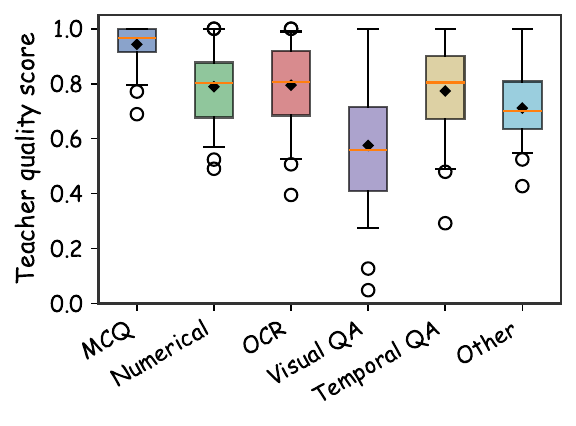}
    \end{minipage}\hfill
    \begin{minipage}[t]{0.48\linewidth}
        \centering
        \includegraphics[width=\linewidth]{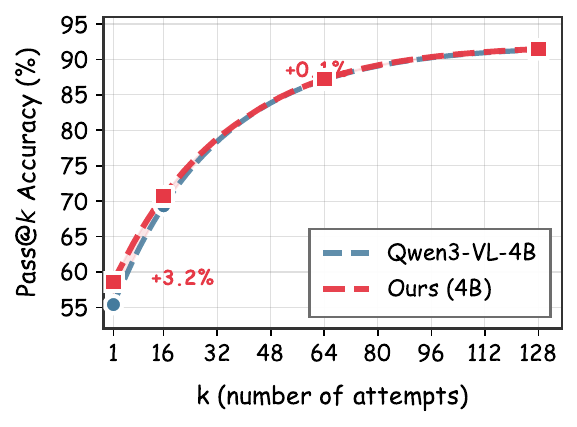}
    \end{minipage}
    \caption{Per-task teacher variance and Pass@$k$ behavior on Video-MMMU.
    \textbf{Left:} Teacher quality distribution by task type; box plots over 200 teacher responses across six task families highlight that high-stability closed-ended tasks (e.g., MCQ, numerical) have lower variance, while higher-variance tasks such as visual QA exhibit wider spreads.
    \textbf{Right:} Pass@$k$ accuracy on Video-MMMU as a function of $k$ (number of sampled student responses per query); our method achieves +3.2\% higher Pass@1 accuracy than Qwen3-VL-4B while converging to a similar upper bound as $k$ increases.}
    \label{fig:variance_and_passk}
\end{figure}

We analyze 200 teacher responses from a frozen Qwen3-VL-235B teacher across 8 task types.
Figure~\ref{fig:variance_and_passk} summarizes per-task quality distributions and Pass@$k$ behavior.

Cross-question variance is substantial: quality ranges from 0.10 to 1.0 ($\mu{=}0.75$, $\sigma{=}0.22$).
MCQ shows high stability ($\mu{=}0.96$, $\sigma{=}0.10$), while visual QA exhibits higher variance ($\mu{=}0.64$, $\sigma{=}0.24$).
Within-question sampling variance is also significant: $\sigma_{\text{sampling}}$ ranges from $0.07$ (MCQ) to $0.15$ (visual QA), with quality spans of $[0.50, 0.85]$ for OCR and $[0.65, 1.00]$ for numerical tasks.
Format violations occur in 1\% of samples overall (10\% for temporal QA).

These trends explain the gains: high-variance tasks such as Video-MMMU ($+3.2$) benefit most from R-MSD, whereas LongVideoBench shows little improvement due to a frame-count mismatch (training uses 16 frames despite its longer-context demands).

\vspace{-8pt}  

\section{Conclusion}
\label{sec:conclusion}

We study teacher decoding uncertainty as a practical bottleneck in video LVLM
distillation. Through empirical analysis, we identify two variance dimensions:
\textbf{(1)} cross-question variance ($\sigma{=}0.22$ globally) where different
questions yield different teacher quality; and \textbf{(2)} within-question decoding
uncertainty ($\sigma_{\text{sampling}}{\approx}0.10$ on average) where repeated samples of the
same question span quality ranges of $[0.50, 0.85]$ for OCR and $[0.65, 1.00]$ for
numerical tasks. Combined with format violations (1.0\% overall, about 10\% for temporal
QA), this makes single-sample supervision fundamentally unreliable.

To address this, we propose R-MSD (Reliable Multi-Sample Distillation), which
samples a teacher pool per input and applies task-adaptive matching:
for closed-ended tasks, it preferentially pairs student rollouts with
higher-quality teacher responses measured against ground truth; for open-ended
tasks, it uses uniform pairing to avoid brittle lexical bias.
Building on this matching strategy, we first perform SFT warmup and then apply
RL-based adversarial distillation, where an online discriminator provides
distribution-level alignment signals.

Across six video benchmarks and two image QA benchmarks, R-MSD consistently
outperforms single-sample and uniform multi-sample baselines under matched
training budgets, with gains of +1.5 on VideoMME, +3.2 on Video-MMMU, and
+3.6 on MathVerse.
Under the same protocol, an original SFT+RL 4B baseline shows only marginal
improvement, further highlighting the benefit of task-adaptive multi-sample supervision.

\textbf{Summary.}
Our findings show that supervision \emph{selection strategy} is as important as supervision quantity in response-based video distillation.
For closed-ended tasks, quality-biased pairing improves performance by filtering low-quality teacher samples; for open-ended tasks, uniform pairing preserves semantic diversity and avoids brittle lexical penalties.
This task-adaptive paradigm improves distillation reliability without extra reward models or complex infrastructure.

\textbf{Limitations.}
Our closed-ended task quality scoring relies on task-specific ground-truth annotations and may
not directly apply in weakly supervised settings.
For open-ended tasks, uniform weighting is a conservative choice that preserves
semantic diversity but does not explicitly rank semantic correctness.
The multi-sample protocol increases training-time compute approximately
proportional to $K$.

\textbf{Future work.}
Future work includes semantic quality estimators for open-ended responses,
adaptive selection of $K$ based on input difficulty, and integration with other
distillation/alignment objectives.
Reference-aware reward correction in strong-to-weak settings is another
promising direction for improving quality-signal fidelity when teacher and
student capacities differ substantially.

\bibliographystyle{unsrtnat}
\bibliography{main}

\end{document}